\documentclass[runningheads]{llncs}

\usepackage{accv}

\usepackage{accvabbrv}

\usepackage[T1]{fontenc}
\usepackage[utf8]{inputenc}
\usepackage{graphicx}
\usepackage{amsmath,amssymb}
\usepackage{bbm}
\usepackage{booktabs}
\usepackage{multirow}
\usepackage{makecell}
\usepackage{array}
\usepackage{url}
\usepackage{pifont}
\usepackage{xspace}
\usepackage[table]{xcolor}
\usepackage{wrapfig}

\usepackage[accsupp]{axessibility}  

\usepackage{hyperref}

\usepackage{orcidlink}

\newcommand{\method}{APRM\xspace}

\graphicspath{{figs/}{figures/}}

\begin{document}

\title{Anatomical Prior-Guided Routing for 3D Lung Tumor Segmentation} 

\titlerunning{Anatomical Prior-Guided Routing for 3D Segmentation}

\makeatletter
\renewcommand\@fnsymbol[1]{\ensuremath{\ifcase#1\or \star\or \dagger\or
   \ddagger\or \mathsection\or \mathparagraph\or \|\or **\or \dagger\dagger
   \or \ddagger\ddagger \else\@ctrerr\fi}}
\makeatother

\author{Uihyeon Maeng\inst{1}\thanks{Equal Contribution}\orcidlink{0009-0002-2692-7907} \and
Jonghun Kim\inst{2}$^\star$ \orcidlink{0009-0002-2790-2090} \and
Hyunjin Park\inst{1,2}\thanks{Corresponding Author}\orcidlink{0000-0001-5681-8918}}

\authorrunning{U. Maeng et al.}

\institute{Department of Artificial Intelligence, Sungkyunkwan University, Suwon, Korea \\
\and
Department of Electrical and Computer Engineering, \\ Sungkyunkwan University, Suwon, Korea \\
\email{\{uhmaeng,iproj2,hyunjinp\}@skku.edu}}

\maketitle

\begin{abstract}
Accurate lung tumor segmentation in computed tomography (CT) is essential for treatment planning, response assessment, and quantitative imaging analysis. However, 3D lung tumor segmentation remains challenging because tumors occupy only a small portion of the whole volume and exhibit heterogeneous appearances across anatomical locations. In patch-based training and sliding-window inference for 3D medical image segmentation, location-dependent heterogeneity is further amplified because cropped image patches provide limited global anatomical context, making it difficult for a model to adapt its feature processing according to the anatomical region from which each patch is extracted. To address this limitation, we propose an Anatomical Prior-guided Routing Module (APRM) for lung tumor segmentation. For each image patch, we obtain anatomical prior information of regional occupancy and deterministically select an anatomical routing gate. The selected gate then performs sparse feature routing over a shared expert adapter pool, enabling anatomy-specific feature adaptation while sharing expert capacity across regions.
We evaluate APRM on two datasets with various backbone networks and demonstrate consistent improvements over baseline models. Furthermore, feature visualizations show that APRM produces different feature responses depending on patch location, indicating that the proposed routing mechanism effectively adapts feature processing according to anatomical context.
  \keywords{Computed Tomography \and 3D Medical Image Segmentation \and Anatomical Prior-guided Routing}
\end{abstract}

\section{Introduction}
\label{sec:intro}

\begin{figure}[t]
    \centering
    \begin{minipage}[b]{0.42\textwidth}
        \centering
        \includegraphics[width=\textwidth]{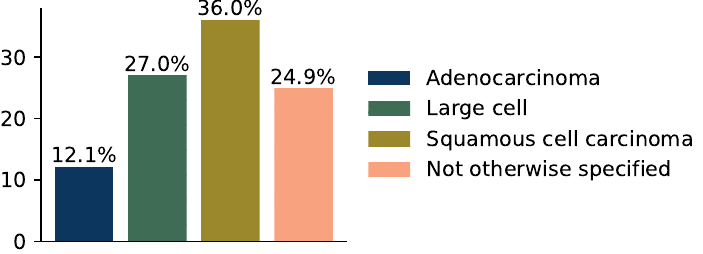}
        \caption*{\scriptsize (a) Subtype Distribution of Lung Tumors}
    \end{minipage}
    \centering
    \begin{minipage}[b]{0.57\textwidth}
        \centering
        \includegraphics[width=\textwidth]{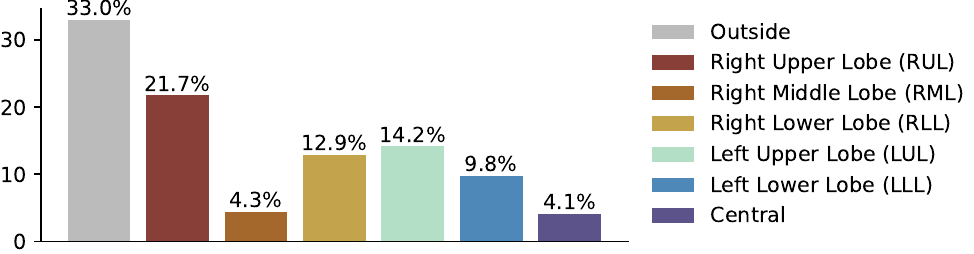}
        \caption*{\scriptsize (b) Anatomical Distribution of Lung Tumors}
    \end{minipage}
    \caption{Distribution of lung cancer subtypes and anatomical tumor locations within the NSCLC\cite{aerts2014decoding,clark2013cancer,prior2017public} dataset.
    Tumor location is defined as the anatomical region (e.g., right upper lobe) showing the maximum spatial overlap with the tumor. Outside indicates areas around the lungs that do not overlap with any anatomical masks. } 
    \label{fig1}
\end{figure}

Lung cancer is one of the most common cancers worldwide \cite{bray2024global} and is characterized by high incidence and mortality rates that demand early detection and precise management~\cite{li2022advances}.
To diagnose and formulate treatment plans, computed tomography (CT) is widely used in clinical practice because it provides a non-invasive way to observe lung tumors and surrounding anatomical structures~\cite{latimer2015lung,ettinger2017non}.
Accurate lung tumor segmentation is essential for determining tumor extent, supporting treatment planning, evaluating response, and enabling quantitative imaging analysis~\cite{primakov2022automated,kashyap2025automated,li2024systematic}.
However, tumor segmentation has traditionally relied on manual annotation by experienced clinicians.
Manual annotation is inherently time-consuming, labor-intensive, and prone to high inter-observer variability, which incurs substantial clinical costs. 
These limitations motivate deep learning-based automatic segmentation, which can provide efficient tumor annotation. 
Despite recent advances, lung tumor segmentation remains challenging~\cite{li2024systematic,liu2021review}. 
Lung tumors often occupy only a small fraction of the entire CT volume. 
In addition, tumor boundaries can be obscured by surrounding anatomy such as the pleura, vessels, bronchi, and mediastinum, leading to location-dependent errors~\cite{mercieca2021challenges,kashyap2025automated}.

\begin{figure}[t]
    \includegraphics[width=\textwidth]{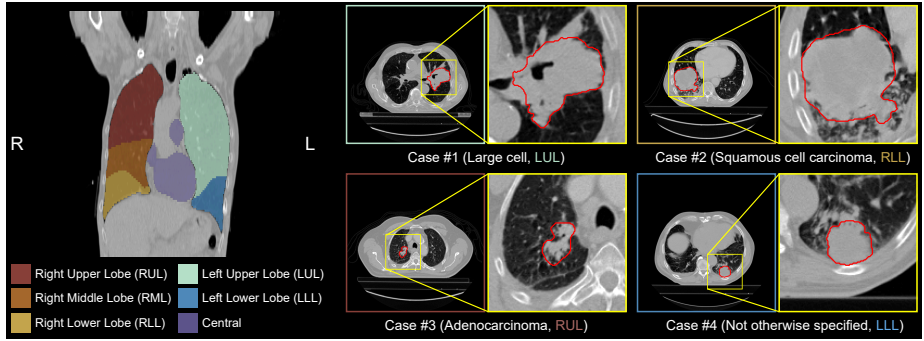}
    \caption{Illustration of lung tumor heterogeneity and anatomical context. Left: anatomical prior regions, including the five lung lobes and the central mediastinum. Right: different histological subtypes and lobar locations. The yellow bounding boxes show tumor regions and the red contours indicate tumor annotations.} 
    \label{fig2}
\end{figure}

Furthermore, lung tumors exhibit heterogeneity across histological subtypes and their imaging appearance can vary with both subtype and anatomical context~\cite{sasaki20242021,lambe2020adenocarcinoma,shi2024radiomics}. As shown in Fig.~\ref{fig1}, both lung cancer subtypes and tumor locations exhibit heterogeneous distributions. In particular, different subtypes appear with different frequencies and lung tumors are not uniformly distributed across anatomical regions. Prior lobe-level analysis has also shown that tumor occurrence and histologic composition differ across anatomical locations, suggesting that tumor subtype and anatomical location are related rather than independent factors~\cite{nilssen2024distribution}. Accordingly, Fig.~\ref{fig2} presents representative cases with different histological subtypes and lobar locations~\cite{aerts2014decoding}. These examples show that tumor-centered patches can contain diverse local anatomical contexts, including lung parenchyma, pleural or chest-wall interfaces, and regions close to cardiomediastinal structures. Such subtype- and location-dependent heterogeneity may influence boundary conspicuity and apparent tumor shape, while introducing nearby non-tumor tissues that can be confused with the tumor.
In addition, medical datasets are often limited in size because data collection and expert annotation are costly~\cite{na2024radiomicsfill,kim2024domain,kim2025tumor,kim2026visual}, making it difficult for conventional segmentation methods to learn the full heterogeneity of lung tumors from image appearance alone. Therefore, incorporating anatomical priors is important: a model should adapt feature processing according to where a tumor patch is located, so that it can better interpret subtype- and location-dependent tumor appearance and distinguish tumors from surrounding anatomical structures~\cite{mercieca2021challenges,kashyap2025automated}.

Mixture-of-experts (MoE) methods perform dynamic routing according to input features, activating only a subset of experts for each input while increasing the overall model capacity~\cite{shazeer2017outrageously,fedus2022switch}. This conditional computation has been used to scale the number of parameters while keeping the number of activated parameters nearly fixed, leading to improved performance and efficiency~\cite{fedus2022switch}. Distribution-aware MoE (dMoE) interprets MoE from a control-theoretic perspective as a feedback control that flexibly adapts to heterogeneous medical image data distributions~\cite{oh2025distributionaware}. However, instead of relying only on feature-driven routing, prior knowledge can provide a meaningful control signal by changing the routing strategy according to the anatomical location.

We propose Anatomical Prior-guided Routing Module (\textbf{APRM}) for 3D lung tumor segmentation. We extract anatomical masks using TotalSegmentator~\cite{wasserthal2023totalsegmentator} and route each image patch based on information from the corresponding anatomical mask. The selected anatomical gate then conditions a router that computes sparse routing weights over a shared expert pool. This design enables anatomical information to guide patch-wise feature adaptation in a stable and interpretable manner, while avoiding the need to allocate a separate expert set to each region. We validate APRM on two datasets with various backbone networks and observe consistent performance improvements when APRM is added to the baseline models. In addition, feature visualizations show that APRM produces different feature responses according to anatomical location, indicating that the proposed module effectively reflects anatomical context during segmentation.

The main contributions of this study are as follows:
\begin{enumerate}
    \item We propose APRM, a plug-in anatomical prior-guided routing module that uses anatomical masks to guide patch-level feature adaptation through gate-conditioned sparse routing.
    \item We validate APRM on two datasets with various backbones, showing consistent performance improvements.
    \item We demonstrate that different feature processing occurs according to the anatomical location, showing the effectiveness of the APRM through feature visualization.
\end{enumerate}

\section{Related Work}

\subsection{Medical Image Segmentation}

Medical image segmentation aims to classify regions of interest, such as lesions, organs, and tumors, at the pixel or voxel level~\cite{jo2024deafusion,kwon2024anatomically,kwon2024enhancing,kwon2024leveraging,kim2025enhancing,kim2025weakly,kwon2025blood,rue2026label}. This task is important for diagnosis, treatment planning, surgical guidance, and quantitative analysis.
U-Net \cite{ronneberger2015u} uses an encoder--decoder structure with skip connections, which is effective for voxel-wise prediction~\cite{ronneberger2015u,cciccek20163d,milletari2016v}. Many U-shaped variants have been proposed to improve this structure. ResUNet adds residual connections to convolution blocks for training deeper networks~\cite{diakogiannis2020resunet}. The nnU-Net instead uses a basic U-Net and automatically adapts preprocessing, training settings, and network configuration to each dataset~\cite{isensee2021nnu}.
Transformer-based methods have also been studied to capture wider context. UNETR uses a transformer encoder with a convolutional decoder~\cite{hatamizadeh2022unetr}, and SwinUNETR uses a hierarchical Swin Transformer~\cite{hatamizadeh2021swin}. Since full-resolution 3D volumes require large memory, these methods usually use patch-based training and sliding-window inference~\cite{isensee2021nnu,jeon2025no}. Recent methods also improve U-Net with modern convolution blocks or large-scale pre-training, such as MedNeXt~\cite{roy2023mednext} and STUNet~\cite{huang2023stu}. Unlike these works, \method does not introduce a new backbone. It is a plug-in routing module that can be added to existing segmentation models.

\subsection{Anatomical Priors and Mixture-of-Experts}

Anatomical priors provide information about body structures and have been used as spatial maps, shape constraints, auxiliary inputs, or post-processing steps in medical image segmentation~\cite{liu2021anatomy,larrazabal2020post,huang2021medical,you2024learning}. Automatic CT parsers such as TotalSegmentator can extract organ masks from CT images~\cite{wasserthal2023totalsegmentator}. These masks can provide anatomical context for tasks such as lymph-node segmentation~\cite{mathai2024segmentation,engelson2024lnq}.
MoE models use conditional computation. A router selects a small number of experts for each token or patch, so the model can increase its capacity without using all experts at once~\cite{shazeer2017outrageously,fedus2022switch,zhou2022mixture,riquelme2021scaling}. Recent medical imaging studies have used MoE for continual learning, multi-modal representation, SAM adaptation, and semi-supervised learning~\cite{chen2024low,jiang2024m4oe,li2025moe}. These methods usually use image features, modalities, or prompts for routing.
However, feature-based routing methods may overlook anatomical location. In lung tumor segmentation, errors often depend on anatomical location because nearby structures such as the pleura, vessels, and mediastinum can obscure tumor boundaries. We use anatomical priors as the routing context. This allows the model to adapt patch-level features according to the location of each patch.

\begin{figure}[t]
\centering
\includegraphics[width=\textwidth]{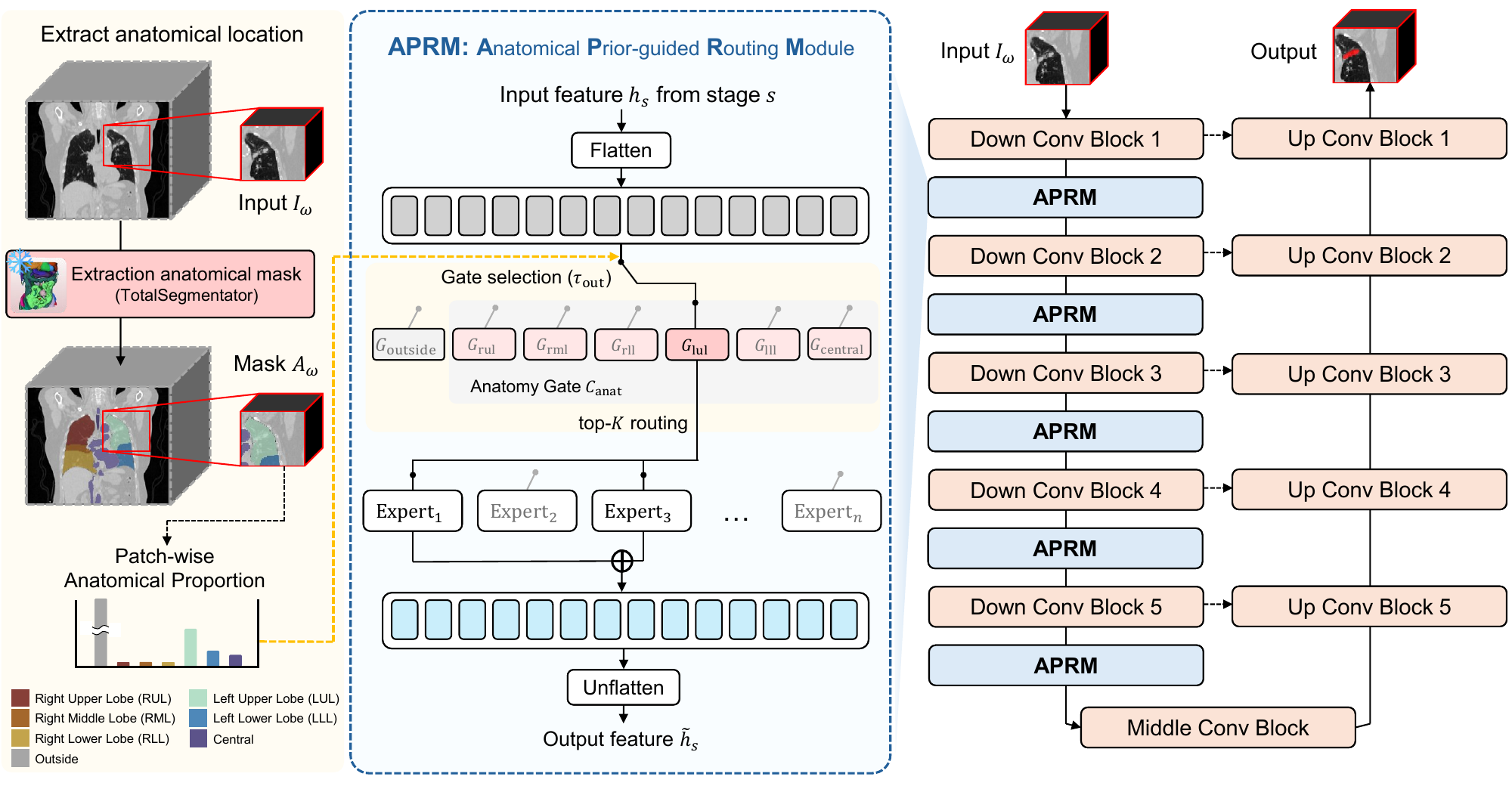}
\caption{Overview of proposed APRM. Anatomical masks are extracted using a pretrained TotalSegmentator \cite{wasserthal2023totalsegmentator}. An anatomical gate is assigned based on these anatomical masks. When the outside region is larger than \(\tau_\text{out}\), the patch is assigned to the outside gate. The selected gate computes top-\(K\) sparse weights using shared experts.}
\label{fig3}
\end{figure}

\section{Method}

We propose APRM, which adapts feature processing according to anatomical locations to reflect tumor heterogeneity.
APRM is added to all scales of the backbone encoder and applies different feature routing strategies according to the anatomical location of each patch.
Fig.~\ref{fig3} provides an overview of APRM. Let input image \(I \in \mathbb{R}^{H \times W \times D}\) denote a CT volume, where \(H\), \(W\), and \(D\) are the spatial dimensions. Let \(\omega \in \mathbb{R}^{h \times w \times d}\) denote a patch. The image is divided into patches. We define the patch set \(\Omega\) as
\begin{equation}
\Omega=\left\{\omega \mid \omega \subset \{1,\ldots,H\}\times\{1,\ldots,W\}\times\{1,\ldots,D\},\ \omega \in \mathbb{R}^{h\times w\times d} \right\},
\end{equation}
where patch \(\omega \in \Omega\) denotes a patch region. The local image patch extracted from a patch region \(\omega\) is denoted by \(I_{\omega}\).
We use an anatomical label map \(A\) extracted using TotalSegmentator~\cite{wasserthal2023totalsegmentator}. We compute the occupancy ratio of anatomical regions in \(I_{\omega}\) from \(A\) for routing gate selection. If the outside proportion exceeds a threshold \(\tau_\text{out}\), the patch is assigned to the outside gate. Otherwise, the anatomical gate is selected from among the five lung lobes and the central region.
The selected gates independently route features to the shared experts. This design allows routing to vary according to anatomical context while enabling expert capacity to be shared across multiple anatomical regions.

\subsection{Definition of Anatomical Prior}
\label{sec:anatomical_prior}

We predefine anatomical regions relevant to lung cancer and extract their masks using TotalSegmentator~\cite{wasserthal2023totalsegmentator}. We use masks for the five lung lobes: right upper lobe (RUL), right middle lobe (RML), right lower lobe (RLL), left upper lobe (LUL), and left lower lobe (LLL). We also define a central region by merging masks of selected mediastinal structures, including the heart, aorta, trachea, esophagus, pulmonary vein, superior vena cava, and inferior vena cava. 
The anatomical regions, therefore, consist of the five lung lobes and the central region. Areas that do not belong to these six predefined regions are defined as outside, resulting in seven classes: 
\begin{equation}
\mathcal{C}=\{c_\mathrm{outside},c_\mathrm{rul},c_\mathrm{rml},c_\mathrm{rll},c_\mathrm{lul},c_\mathrm{lll},c_\mathrm{central}\}, \quad 
\mathcal{C}_{\mathrm{anat}}=\mathcal{C}\setminus\{c_\mathrm{outside}\},
\label{eq:class_set}
\end{equation}
where \(\mathcal{C}\) denotes the set of anatomical region classes and \(\mathcal{C}_{\mathrm{anat}}\) denotes the subset excluding the outside region. 

\subsection{Anatomical Prior-Guided Routing}
\label{sec:aprm}

\textbf{Gate selection.}
We obtain the cropped anatomical label map \(A_{\omega}\) corresponding to the image patch \(I_{\omega}\). To determine which anatomical gate should be used for the patch, we calculate the occupancy ratio of each predefined mask in the patch. The anatomical occupancy ratio is defined as
\begin{equation}
q(\omega)=\frac{1}{|\omega|}\sum_{u \in A_\omega}\mathbbm{1}\{A_\omega(u)=c\},\quad c \in \mathcal{C},
\end{equation}
where \(|\omega|\) denotes the total number of voxels in the patch, the numerator counts the voxels in the cropped anatomical label map \(A_{\omega}\) that belong to class \(c\), and \(\mathbbm{1}\) is an indicator function. 
The outside occupancy \(q_\mathrm{outside}(\omega)\) is computed to classify the patch \(\omega\) into outside or anatomical regions. If this value is greater than or equal to the threshold \(\tau_{\mathrm{out}}\), the patch is defined as outside. If the outside occupancy is smaller than the threshold, the patch is assigned to the class with the largest occupancy among \(\mathcal{C}_{\mathrm{anat}}\):
\begin{equation}
G_{\omega}=
\begin{cases}
    G_\mathrm{outside}, & q_{\mathrm{outside}}(\omega)\geq \tau_{\mathrm{out}},\\
    G_{c^*},\ c^*=\operatorname*{arg\,max}_{c \in \mathcal{C}_\mathrm{anat}} q_c(\omega), & q_{\mathrm{outside}}(\omega)<\tau_{\mathrm{out}},
\end{cases}
\label{eq:gate_selection}
\end{equation}
where \(G_{\omega}\) denotes the routing gate for patch \(\omega\). According to the anatomical location classes in Eq.~\ref{eq:class_set}, a routing gate is assigned to each anatomical region.

\vspace{3pt}
\noindent \textbf{Feature routing with anatomical gates.}
APRM is applied at each encoder stage and the anatomical gate is selected according to the anatomical location of each patch.
The selected gate routes the feature tokens to a shared expert pool, allowing the routing function to be anatomy-specific while keeping the expert capacity shared across all anatomical regions.
For expert routing, the feature map \(h_s\) at stage \(s\) is flattened into a spatial token sequence \(X_s\):
\begin{equation}
X_s=\operatorname{Flatten}(h_s)\in\mathbb{R}^{N_s\times C_s},\quad h_s\in\mathbb{R}^{C_s\times H_s\times W_s \times D_s},\quad N_s = H_s W_s D_s,
\end{equation}
where \(C_s\), \(H_s\), \(W_s\), and \(D_s\) denote the channel, height, width, and depth at stage \(s\), respectively.
After flattening, the feature map is represented by \(N_s\) tokens, each with a dimension \(C_s\).
For each token \(x_s \in X_s\), the assigned gate \(G_{\omega}\) computes sparse expert weights over shared experts and the routed output is obtained as
\begin{equation}
    \tilde{x}_s = x_s + \sum_{i=1}^{M}G_\omega(x_s)_i\cdot E_i(x_s),
\end{equation}
where \(M\) is the number of shared experts, \(E_i(x_s)\) denotes the output of the \(i\)-th expert network, and \(G_\omega(x_s)\) is the sparse routing weight assigned by gate \(G_\omega\). 
To prevent the model from relying on only a small subset of experts, we employ noisy top-\(K\) routing, which injects Gaussian noise into the routing logits before selecting the active experts. The noisy logits and sparse gating weights are defined as follows:
\begin{equation}
    G_\omega(x_s)=\operatorname{Softmax}\left(\operatorname{KeepTopK}(H(x_s), K)\right),
\end{equation}
\begin{equation}
    H(x_s)_i = (x_s W_g)_i + \mathcal{N}(0,1)\cdot \operatorname{Softplus}\left((x_s W_{\mathrm{noise}})_i\right),
\end{equation}
\begin{equation}
    \operatorname{KeepTopK}(v, K)_i = \begin{cases}
v_i & \text{if } v_i \text{ is among the top-}K\text{ elements of } v, \\
-\infty & \text{otherwise.}
\end{cases}
\end{equation}
where \(W_g\) and \(W_{\mathrm{noise}}\) are the learnable parameters of the selected anatomical gate for routing logits and noise scale, respectively. The \(\operatorname{Softplus}\) function constrains the noise scale to be positive.
After all tokens are routed, the updated token sequence \(\tilde{X}_s=\{\tilde{x}_s\}_{n=1}^{N_s}\) is reshaped back to the feature-map resolution:
\begin{equation}
\tilde{h}_s=\operatorname{Unflatten}(\tilde{X}_s)\in\mathbb{R}^{C_s \times H_s \times W_s \times D_s},
\end{equation}
where \(\tilde{h}_s\) is the output feature map at stage \(s\).

\subsection{Objective Function}
The segmentation objective is defined as the sum of Dice loss and cross-entropy (CE) loss:
\begin{equation}
\mathcal{L}_{\mathrm{seg}}=\mathcal{L}_{\mathrm{Dice}}+\mathcal{L}_{\mathrm{CE}}.
\end{equation}
In sparse routing, the router may repeatedly assign high probabilities to only a small subset of experts, which can reduce expert diversity and limit the effective capacity of the shared expert pool. To mitigate this issue, we introduce a lightweight expert balancing loss \cite{fedus2022switch}. This loss is computed from the dense softmax probabilities of the selected router before top-\(K\) masking, so that all experts receive a learning signal regardless of whether they are selected by the sparse routing operation.
\begin{equation}
p(x_s)_i=\frac{\exp\left(H(x_s)_i\right)}{\sum_{j=1}^{M}\exp\left(H(x_s)_j\right)}.
\end{equation}
Using these dense routing probabilities, the average usage probability of expert \(i\) at stage \(s\) is computed over all tokens in the stage as
\begin{equation}
\bar{p}(X_s)_i=\frac{1}{N_s}\sum_{x_s \in X_s}p(x_s)_i.
\end{equation}
The expert balancing loss is then defined by penalizing concentrated average expert usage across the routed encoder stages:
\begin{equation}
\mathcal{L}_{\mathrm{bal}}=\frac{1}{|\mathcal{S}_r|}\sum_{s \in \mathcal{S}_r}M\sum_{i=1}^{M}\left(\bar{p}(X_s)_i\right)^2,
\end{equation}
where \(\mathcal{S}_r\) denotes the set of encoder stages. By minimizing this term, the model is encouraged to distribute the average dense routing probability more evenly across the shared expert pool rather than concentrating on a few experts. The final training objective combines the segmentation loss with the balancing loss as
\begin{equation}
\mathcal{L}=\mathcal{L}_{\mathrm{seg}}+\lambda\mathcal{L}_{\mathrm{bal}}.
\label{eq:loss}
\end{equation}
The balancing weight \(\lambda\) is empirically set to \(10^{-2}\) in our experiments.

\subsection{Sampling Process}
During training, a cropped region (i.e., patch) \(\omega\) is sampled from the input volume \(I\), and the corresponding image patch \(I_{\omega}\) and anatomical patch \(A_{\omega}\) are extracted together. The anatomical patch is used to determine the routing gate according to Eq.~\eqref{eq:gate_selection}. The anatomical prior used for routing is spatially aligned with the feature tokens processed by APRM at each encoder stage.
During sampling, we process the full volume using sliding-window inference with an overlap ratio of 0.5 along each spatial dimension. Predictions from overlapping windows are combined using Gaussian-weighted voxel-wise aggregation, following the standard nnU-Net inference procedure. For each patch, the same crop is applied to the anatomical label map and the patch-level anatomical occupancy is computed to select the corresponding anatomical gate. Predictions from overlapping windows are aggregated in the original image space and the final segmentation is obtained from the merged probability map. 

\section{Experiments}

\subsection{Experimental Setup}
\label{sec:experimental_setup}

\textbf{Datasets.}
We evaluate 3D lung tumor segmentation using two public CT datasets: NSCLC~\cite{aerts2014decoding,clark2013cancer,prior2017public} and MSD~\cite{antonelli2022medical}. Both NSCLC and MSD are non-small cell lung cancer datasets. NSCLC provides subtype annotations consisting of large cell carcinoma, squamous cell carcinoma (SCC), adenocarcinoma (Aden.), and not otherwise specified (NOS), whereas MSD does not provide subtype information. Following the nnU-Net preprocessing setup, both datasets were resampled to a target spacing of \(1\times1\times3\,\mathrm{mm}^3\) and have an image size of \(512\times512\times(82\text{--}176)\). NSCLC is used for model training and evaluation. MSD is used as an external test dataset. From NSCLC, we use 319 subjects for training and 80 subjects for test. From MSD, we use 63 subjects for test. 

\vspace{3pt}
\noindent \textbf{Implementation details.}
All methods are implemented using the nnU-Net framework~\cite{isensee2021nnu,isensee2024nnu}. 
The input shape of $224\times192\times48$ was automatically determined during the nnU-Net preprocessing steps.
We use the training objective in Eq.~\eqref{eq:loss}, with the balancing loss weight set to \(\lambda=10^{-2}\).
Anatomical masks are extracted using TotalSegmentator~\cite{wasserthal2023totalsegmentator}.
We use seven anatomical routing gates and a shared expert pool per routed stage with \(M=8\) experts and Top-$K$ parameter set to \(K=2\). 
For the +MoE, we use the same expert architecture, number of experts \(M=8\), top-\(K\) sparsity \(K=2\), and sparse weighted aggregation as APRM, but use a single feature-driven router.

\vspace{3pt}
\noindent \textbf{Evaluation metrics.}
We use Dice similarity coefficient (DSC) and intersection over union (IoU) to evaluate volumetric segmentation performance. 

\vspace{3pt}
\noindent \textbf{Comparison backbones.}
We compare four backbone architectures: the U-Net~\cite{ronneberger2015u} architecture in nnU-Net~\cite{isensee2021nnu}, ResEnc~\cite{isensee2024nnu}, MedNeXt~\cite{roy2023mednext}, and STUNet~\cite{huang2023stu}. All backbones are trained and evaluated within the nnU-Net framework.

\begin{table*}[t]
    \centering
    \caption{Quantitative lung tumor segmentation results on the NSCLC~\cite{aerts2014decoding,clark2013cancer,prior2017public} dataset with various backbones. Results are reported for each tumor subtype, where SCC denotes squamous cell carcinoma, Aden. denotes adenocarcinoma, and NOS denotes not otherwise specified.
    Average indicates the mean performance across all subjects. 
    DSC and IoU are reported as percentages and higher values indicate better performance. Bold indicates the \textbf{best} performance.}
    \label{tab:nsclc}
    \scriptsize
    \setlength{\tabcolsep}{3pt}
    \renewcommand{\arraystretch}{1.2}
    \resizebox{\textwidth}{!}{%
        \begin{tabular}{llcccccccccc}
            \toprule
             & Dataset & \multicolumn{10}{c}{NSCLC~\cite{aerts2014decoding,clark2013cancer,prior2017public}}  \\
            \cmidrule(lr){3-12}
            & Subtype & \multicolumn{2}{c}{Large cell} & \multicolumn{2}{c}{SCC} & \multicolumn{2}{c}{Aden.} & \multicolumn{2}{c}{NOS} & \multicolumn{2}{c}{Average}  \\
            \cmidrule(lr){3-4} \cmidrule(lr){5-6} \cmidrule(lr){7-8} \cmidrule(lr){9-10} \cmidrule(lr){11-12}  
            Backbone & Module & DSC(\%) & IoU(\%) & DSC(\%) & IoU(\%) & DSC(\%) & IoU(\%) & DSC(\%) & IoU(\%) & DSC(\%) & IoU(\%) \\
            \midrule
            \multirow{3}{*}{nnU-Net~\cite{isensee2021nnu}}
             & Baseline & 71.30 & 61.28 & 59.18 & 48.05 & \textbf{79.66} & \textbf{70.70} & 57.66 & 48.27 & 65.64 & 55.64   \\
             & + MoE & \textbf{72.52} & \textbf{61.86} & \textbf{62.54} & \textbf{51.10} & 77.77 & 68.79 & 57.65 & 48.95 & 66.66 & 56.55  \\
             & \cellcolor{gray!15}+ \method & \cellcolor{gray!15}72.38 & \cellcolor{gray!15}61.39 & \cellcolor{gray!15}62.38 & \cellcolor{gray!15}50.93 & \cellcolor{gray!15}77.30 & \cellcolor{gray!15}68.05 & \cellcolor{gray!15}\textbf{60.62} & \cellcolor{gray!15}\textbf{50.90} & \cellcolor{gray!15}\textbf{67.29} & \cellcolor{gray!15}\textbf{56.75}  \\
            \midrule
            \multirow{3}{*}{ResEnc~\cite{isensee2024nnu}}
             & Baseline & 73.54 & 62.32 & 63.37 & 51.41 & 84.91 & 75.40 & 64.22 & 54.27 & 70.00 & 59.17 \\
             & + MoE & 73.27 & 61.83 & \textbf{64.10} & \textbf{52.53} & 83.48 & 73.94 & 63.20 & 53.65 & 69.63 & 58.94 \\
             & \cellcolor{gray!15}+ \method & \cellcolor{gray!15}\textbf{73.63} & \cellcolor{gray!15}\textbf{62.44} & \cellcolor{gray!15}63.38 & \cellcolor{gray!15}51.69 & \cellcolor{gray!15}\textbf{86.88} & \cellcolor{gray!15}\textbf{77.53} & \cellcolor{gray!15}\textbf{66.32} & \cellcolor{gray!15}\textbf{56.17} & \cellcolor{gray!15}\textbf{70.88} & \cellcolor{gray!15}\textbf{60.10} \\
            \midrule
            \multirow{3}{*}{MedNeXt~\cite{roy2023mednext}}
             & Baseline & 66.80 & 56.06 & 54.63 & 43.20 & 70.84 & 59.61 & 57.22 & 47.62 & 64.08 & 53.13 \\
             & + MoE & \textbf{72.76} & \textbf{61.91} & 63.04 & 51.36 & 78.70 & 69.60 & \textbf{60.54} & \textbf{51.03} & 67.76 & 57.30 \\
             & \cellcolor{gray!15}+ \method & \cellcolor{gray!15}72.52 & \cellcolor{gray!15}61.43 & \cellcolor{gray!15}\textbf{65.53} & \cellcolor{gray!15}\textbf{53.54} & \cellcolor{gray!15}\textbf{85.37} & \cellcolor{gray!15}\textbf{75.20} & \cellcolor{gray!15}58.38 & \cellcolor{gray!15}49.37 & \cellcolor{gray!15}\textbf{68.81} & \cellcolor{gray!15}\textbf{58.16} \\
             \midrule
             \multirow{3}{*}{STUNet~\cite{huang2023stu}}
             & Baseline & 72.44 & 61.37 & 57.00 & 46.68 & 75.46 & 65.73 & 58.75 & 49.56 & 65.05 & 54.88 \\
             & + MoE & 66.68 & 56.47 & \textbf{62.29} & \textbf{51.15} & 76.12 & 66.04 & 60.54 & \textbf{51.43} & 65.28 & 55.06 \\
             & \cellcolor{gray!15}+ \method & \cellcolor{gray!15}\textbf{73.27} & \cellcolor{gray!15}\textbf{62.30} & \cellcolor{gray!15}56.48 & \cellcolor{gray!15}46.12 & \cellcolor{gray!15}\textbf{79.67} & \cellcolor{gray!15}\textbf{69.41} & \cellcolor{gray!15}\textbf{60.67} & \cellcolor{gray!15}50.68 & \cellcolor{gray!15}\textbf{66.31} & \cellcolor{gray!15}\textbf{55.87} \\
            \bottomrule
        \end{tabular}%
    }
\end{table*}

\subsection{Results of Lung Tumor Segmentation}
\label{sec:main_results}

\begin{figure}[t]
\centering
\includegraphics[width=\linewidth]{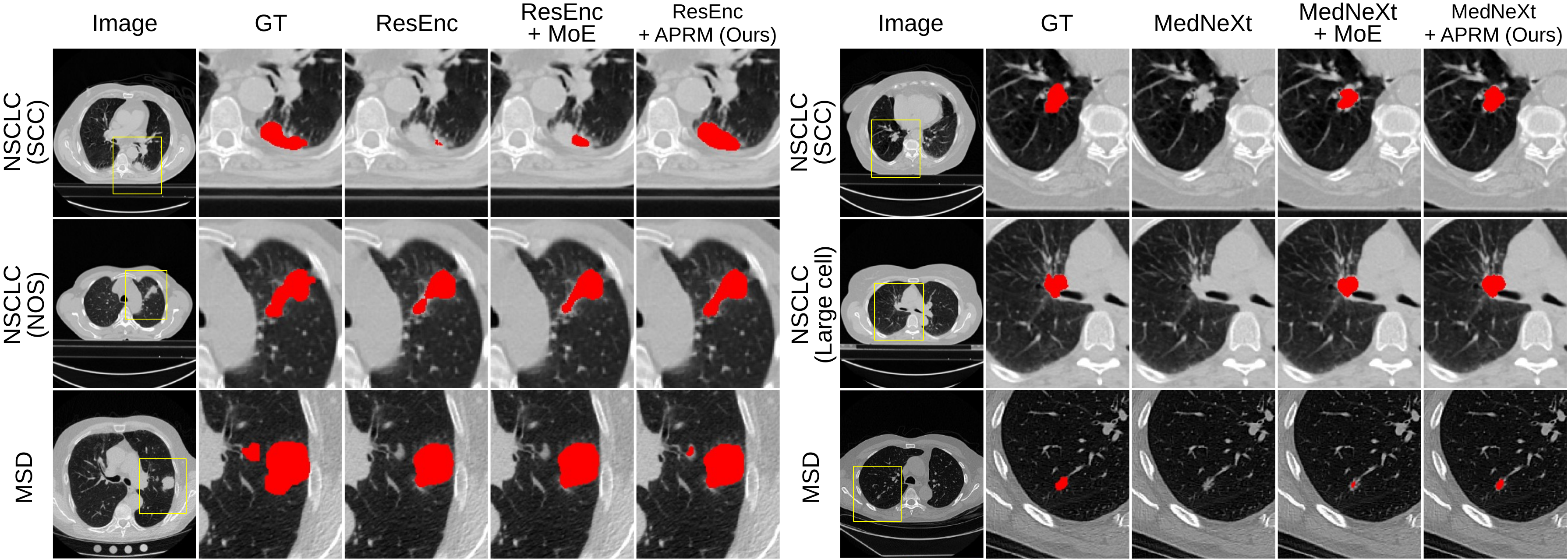}
\caption{Qualitative lung tumor segmentation results. Left: comparison with different modules using ResEnc as the backbone. Right: comparison with different modules using MedNeXt as the backbone. The top two rows show results for different NSCLC tumor subtypes, while the bottom row shows results on the MSD dataset. Each result shows a zoomed-in view of the yellow bounding box in the image.}
\label{fig4}
\end{figure}

\noindent \textbf{Results on NSCLC.}
We compare APRM with the baseline and MoE across several backbones, as shown in Table~\ref{tab:nsclc}. 
Among the backbones, ResEnc achieves the best performance.
Adding MoE improves the results for most tumor subtypes, while Adding APRM improves the performance of all backbones and achieves the best average scores for every backbone. 
These results indicate that anatomy-guided routing enables effective feature adaptation across heterogeneous tumor groups.
Fig.~\ref{fig4} (top) shows qualitative results on NSCLC, where the tumors have different sizes and locations depending on the subtype. 
The baseline often fails to segment the tumor region and MoE captures part of the tumor but misses fine details. 
In contrast, APRM produces results that are closest to the ground truth. 
These results show that APRM can segment tumors well across different locations and sizes, demonstrating the effectiveness of the proposed module.

\begin{wraptable}{r}{0.5\textwidth}
    \vspace{-38pt}
    \centering
    \captionof{table}{Quantitative lung tumor segmentation results on the MSD dataset~\cite{antonelli2022medical}. Bold indicates the \textbf{best} performance.}
    \vspace{6pt}
    \label{tab:msd}
    \scriptsize
    \setlength{\tabcolsep}{3pt}
    \renewcommand{\arraystretch}{1.2}
    \scalebox{0.98}{%
        \begin{tabular}{llcc}
            \toprule
             & Dataset & \multicolumn{2}{c}{MSD~\cite{antonelli2022medical}}  \\
            \cmidrule(lr){3-4}
            Backbone & Module & DSC(\%) & IoU(\%) \\
            \midrule
            \multirow{3}{*}{nnU-Net~\cite{isensee2021nnu}}
             & Baseline & 54.86 & 43.09 \\
             & + MoE & 55.18 & 43.11 \\
             & \cellcolor{gray!15}+ \method & \cellcolor{gray!15}\textbf{58.41} & \cellcolor{gray!15}\textbf{46.32}  \\
            \midrule
            \multirow{3}{*}{ResEnc~\cite{isensee2024nnu}}
             & Baseline & 66.38 & 53.47 \\
             & + MoE & 65.91 & 53.30 \\
             & \cellcolor{gray!15}+ \method & \cellcolor{gray!15}\textbf{66.72} & \cellcolor{gray!15}\textbf{54.02}  \\
            \midrule
            \multirow{3}{*}{MedNeXt~\cite{roy2023mednext}}
             & Baseline & 49.36 & 37.69 \\
             & + MoE & 62.49 & 49.96 \\
             & \cellcolor{gray!15}+ \method & \cellcolor{gray!15}\textbf{62.91} & \cellcolor{gray!15}\textbf{50.33} \\
             \midrule
             \multirow{3}{*}{STUNet~\cite{huang2023stu}}
             & Baseline & 56.28 & 43.99\\
             & + MoE & 55.94 & 43.83 \\
             & \cellcolor{gray!15}+ \method & \cellcolor{gray!15}\textbf{58.93} & \cellcolor{gray!15}\textbf{46.83}\\
            \bottomrule
        \end{tabular}%
    }
    \vspace{-27pt}
\end{wraptable}

\vspace{3pt}
\noindent \textbf{Results on MSD.}
We also evaluate each baseline and our module on MSD, which is used as an external test dataset. The quantitative results are shown in Table~\ref{tab:msd}. 
Consistent with the results on NSCLC, ResEnc achieves the best performance among the baseline models, while APRM improves segmentation performance across all evaluated backbones.
The bottom row examples in Fig.~\ref{fig4} show qualitative results on MSD using ResEnc and MedNeXt for two cases: a large tumor on the left and a small tumor on the right. 
In the left case, the comparison methods miss a region that is separated from the main large tumor, whereas APRM correctly detects it as part of the tumor. 
In the right case, the tumor is very small, and the comparison methods fail to detect it, while APRM successfully captures the tumor region. 

\subsection{Ablation Studies}
\label{sec:ablation}

\begin{table*}[t]
    \centering
    \caption{Ablation study on the NSCLC and MSD. ResEnc is used as the backbone.}
    \label{tab:ablation_design}
    \scriptsize
    \setlength{\tabcolsep}{8pt}
    \renewcommand{\arraystretch}{1.2}
    \begin{tabular}{llcccc}
        \toprule
        & Dataset & \multicolumn{2}{c}{NSCLC~\cite{aerts2014decoding,clark2013cancer,prior2017public}}  & \multicolumn{2}{c}{MSD~\cite{antonelli2022medical}} \\
        \cmidrule(lr){3-4} \cmidrule(lr){5-6}
        Category & Setting & DSC(\%) & IoU(\%) & DSC(\%) & IoU(\%) \\
        \midrule
        \multirow{2}{*}{Top-$K$ sparsity}                         & $K=1$      & 70.54 & 59.93 & 65.06 & 52.36 \\
                                                        & \cellcolor{gray!15}$K=2$
                                                        & \cellcolor{gray!15}\textbf{70.88} 
                                                        & \cellcolor{gray!15}\textbf{60.10} 
                                                        & \cellcolor{gray!15}\textbf{66.72}
                                                        & \cellcolor{gray!15}\textbf{54.02} \\
        \midrule
        \multirow{2}{*}{Number of experts}                      & $M=4$      & 70.80 & \textbf{60.21} & 65.74 & 53.29  \\
                                                        & \cellcolor{gray!15}$M=8$ 
                                                        & \cellcolor{gray!15}\textbf{70.88} 
                                                        & \cellcolor{gray!15}60.10 
                                                        & \cellcolor{gray!15}\textbf{66.72}
                                                        & \cellcolor{gray!15}\textbf{54.02} \\
        \midrule
                                                        & $\tau_\mathrm{out}=0.85$ & 69.83 & 59.06 & 66.56 & 53.91 \\
        \rowcolor{gray!15} \cellcolor{white}Outside threshold  & $\tau_\mathrm{out}=0.9$  & \textbf{70.88} & \textbf{60.10} & \textbf{66.72} & \textbf{54.02} \\
                                                        & $\tau_\mathrm{out}=0.95$ & 70.32 & 59.61 & 66.15 & 53.39  \\
        \bottomrule
    \end{tabular}
\end{table*}

\noindent \textbf{Top-$K$ sparsity.}
We observe how the top-$K$ setting affects the sparse routing strategy, as shown in Table~\ref{tab:ablation_design} (top). 
Using the top two experts to aggregate features improves segmentation performance on both NSCLC and MSD compared with using only the top expert. 
Based on this result, we set $K=2$ in our experiments. 

\vspace{3pt}
\noindent \textbf{Number of experts.}
We compare different values of the number of experts \(M\) in the shared expert pool, as shown in Table~\ref{tab:ablation_design} (middle).
Both \(M=4\) and \(M=8\) achieve similar performance, with \(M=8\) providing a slight improvement.
This result suggests that performance does not simply increase by adding more experts. Rather, the routing mechanism dynamically selects the features needed for each input from the shared expert pool.
Based on these results, we use \(M=8\) in our experiments.

\vspace{3pt}
\noindent \textbf{Outside threshold.}
We evaluate different values of the outside threshold \(\tau_\mathrm{out}\) and observe their effect on segmentation performance, as shown in Table~\ref{tab:ablation_design} (bottom).
Because most voxels within a patch are background on average and only a limited portion corresponds to lung-related anatomy, a relatively high \(\tau_\mathrm{out}\) is needed to reliably determine the anatomical location of each patch.
At the lower value \(\tau_\mathrm{out}=0.85\), performance decreases noticeably, indicating that many patches are assigned to the outside category and the model consequently relies predominantly on the outside gate.
This behavior makes the performance close to that of ResEnc + MoE in Tables~\ref{tab:nsclc} and~\ref{tab:msd}.
In contrast, at the higher value \(\tau_\mathrm{out}=0.95\), even a small overlap with an anatomical mask can determine the patch location, which may make anatomical distinctions ambiguous and lead to degraded performance.
We obtain the best performance with \(\tau_\mathrm{out}=0.9\).

\vspace{3pt}

\subsection{Computational Cost Analysis}
\label{sec:efficiency}

\begin{wraptable}{r}{0.6\textwidth}
    \vspace{-32pt}
    \centering
    \caption{Computational cost on the NSCLC test set using nnU-Net. Inference time reports segmentation processing only.}
    \vspace{6pt}
    \label{tab:efficiency}
    \scriptsize
    \setlength{\tabcolsep}{3pt}
    \begin{tabular}{lccc}
        \toprule
        Method & Total params & Router params & Infer. time (s) \\
        \midrule
        nnU-Net & 30.70M & -- & $22.14 \pm 4.80$ \\
        + MoE & 49.38M & 0.017M & $48.57 \pm 15.01$ \\
        + APRM & 49.48M & 0.122M & $50.41 \pm 15.46$ \\
        \bottomrule
    \end{tabular}
    \vspace{-18pt}
\end{wraptable}

We report computational cost using nnU-Net as the backbone.
As shown in Table~\ref{tab:efficiency}, although APRM adds additional anatomy-specific router parameters from 0.017M to 0.122M due to the seven anatomy-specific gates, the total parameter count increases only from 49.38M to 49.48M ($\sim$0.2\%) compared with MoE. Importantly, for each patch, APRM evaluates only the router associated with the selected anatomical gate rather than densely executing all seven routers. Accordingly, segmentation inference increases only modestly from 48.57s to 50.41s ($\sim$3.8\%).

\begin{figure}[t]
\centering
\includegraphics[width=\linewidth]{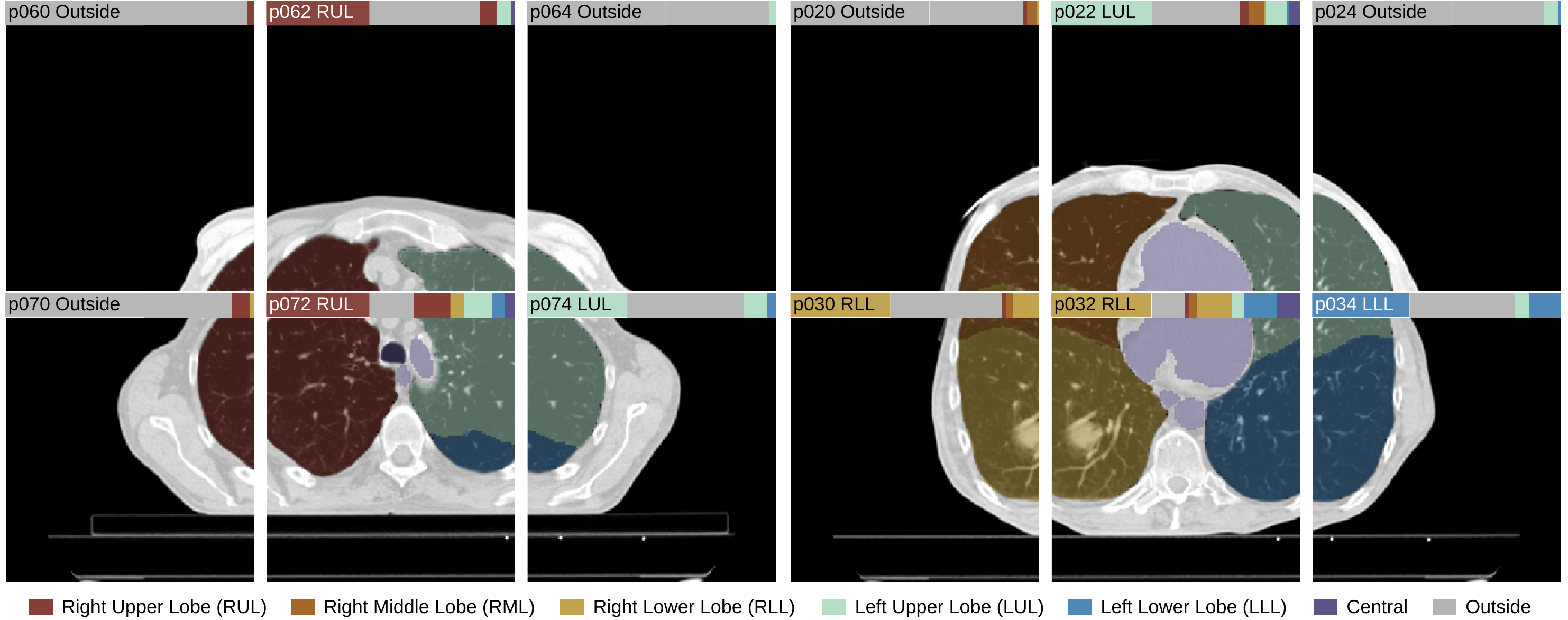}
\caption{Visualization of anatomical patch selection for anatomical gate assignment. Left: an upper (superior) slice closer to the head. Right: a lower (inferior) slice closer to the abdomen. In each patch image, the upper-left label indicates the selected anatomical region, while the upper-right inset shows the distribution of anatomical masks.}
\label{fig5}
\end{figure}

\begin{figure}[t]
\centering
\includegraphics[width=\textwidth]{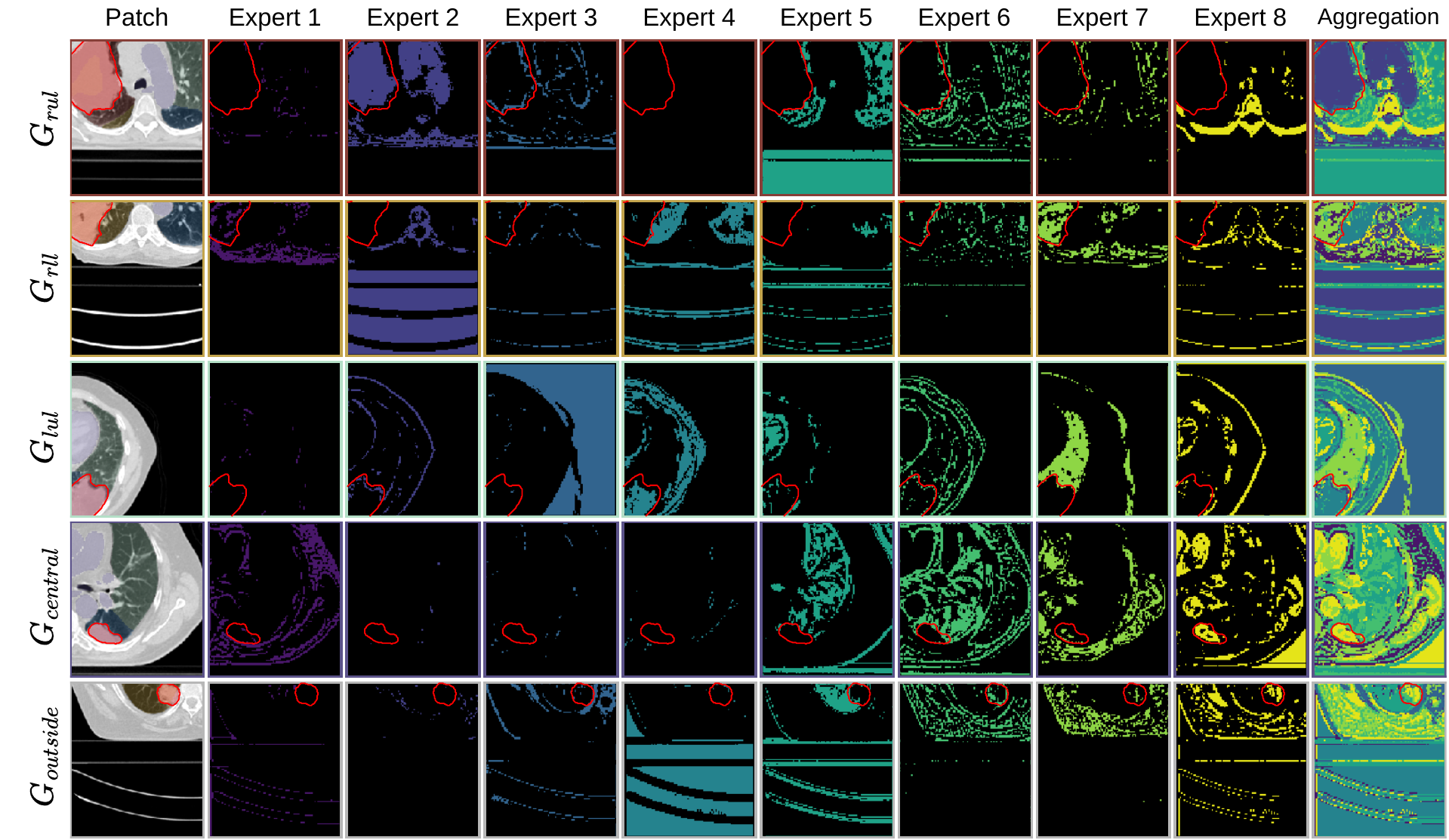}
\caption{Visualization of expert features according to anatomical location. Each image shows the voxels for which the corresponding expert is selected as the top-1 expert and the red contour denotes the tumor region. The aggregation map represents the final feature map obtained after sparse routing.}
\label{fig6}
\end{figure}

\subsection{Analysis of Anatomical Prior-guided Routing}

\vspace{3pt}
\noindent \textbf{Anatomical patch selection.}
Fig.~\ref{fig5} visualizes how patch locations are identified and how anatomical gates are selected during inference.
In the upper-left patch of Fig.~\ref{fig5} (left), only a small portion overlaps with the right upper lobe; therefore, the patch is assigned to the outside region according to \(\tau_\mathrm{out}\).
In the next patch, both the right upper lobe and left upper lobe are present, and the background ratio is below \(\tau_\mathrm{out}\). Therefore, the right upper lobe is selected because it occupies the largest anatomical mask area.
Although the exact tumor location is not explicitly given to the model, sliding-window inference allows each patch to be assigned to different anatomical regions and processed through the corresponding anatomical gates.
This enables the model to select segmentation-relevant features from multiple anatomical contexts and dynamically reflect heterogeneity caused by tumor location, size, and subtype. A similar trend is observed in Fig.~\ref{fig5} (right).

\vspace{3pt}
\noindent \textbf{Routing feature visualization.}
We further visualize the features represented by the shared experts, as shown in Fig.~\ref{fig6}.
Because different experts are assigned at the voxel level, each expert produces a distinct top-1 voxel map.
For the right upper lobe gate (\(G_\mathrm{rul}\), top row), tumor-related regions are mainly selected by Experts 2 and 3: Expert 2 focuses on the tumor interior, while Expert 3 responds to the tumor boundary.
The aggregation feature map preserves these complementary tumor-related responses.
For the right lower lobe gate (\(G_\mathrm{rll}\), second row), tumor-related regions are represented by Experts 1 and 7, showing a routing pattern that differs from that of the right upper lobe gate, and their aggregated feature map also clearly highlights the tumor region.
Similarly, for the left upper lobe gate (\(G_\mathrm{lul}\), third row), Experts 4 and 5 are selected for tumor-related regions.
The central and outside gates also select different experts to represent tumor regions.
These observations show that, although all anatomical gates share the same expert pool, different experts are assigned depending on the anatomical context and tumor appearance.
Thus, the proposed model learns gate-specific routing strategies within the shared expert pool, enabling it to capture heterogeneity in tumor location and size and extract effective segmentation features.

\section{Discussion}
We proposed APRM to incorporate anatomical priors into patch-based 3D lung tumor segmentation. By selecting an anatomical gate for each patch and routing features through a shared expert pool, APRM enables location-aware feature adaptation without assigning separate experts to each anatomical region.
The experimental results show that APRM consistently improves segmentation performance across multiple backbones on the NSCLC dataset and also generalizes to the external MSD dataset. Compared with a standard MoE module, APRM provides more stable gains, indicating that anatomical context is useful beyond simply increasing model capacity. The ablation and visualization results further show that sparse routing with shared experts learns different feature responses for different anatomical regions, helping the model capture tumor interiors, boundaries, and location-dependent heterogeneity.

\vspace{3pt}
\noindent \textbf{Limitations.}
The proposed method depends on anatomical masks extracted by a pretrained segmentation model. Although this makes APRM practical because it does not require manual anatomical annotation, errors in the anatomical masks may lead to incorrect gate assignment, especially near lobe boundaries, mediastinal structures, or abnormal anatomy. 
The anatomical categories used in this work are defined at a relatively coarse level, including lung lobes, the central region, and an outside region. Finer anatomical subdivisions or vessel- and airway-aware priors may provide additional context for tumors with complex local interactions, but they may also increase routing ambiguity and computational cost.
The current framework uses deterministic patch-level gate selection based on anatomical occupancy. This improves stability and interpretability, but it may be suboptimal for patches containing multiple relevant anatomical structures. Future work could explore soft anatomical routing, uncertainty-aware gate assignment, or joint optimization of anatomical priors and tumor segmentation.

\section*{Acknowledgement}
This study was supported by the SMC-SKKU Future Convergence Research Program Grant (SMO126115). It was also supported by the AI Graduate School Support Program (Sungkyunkwan University) (RS-2019-II190421), the National Research Foundation of Korea (RS-2024-00408040, RS-2026-25598771), the ICT Creative Consilience Program Grant (IITP-2026-RS-2020-II201821), the Institute of Information \& Communications Technology Planning \& Evaluation (IITP) grant (RS-2026-25528384, Resource-Intensive AI Technologies Based on Sustainable GPU Integrated Platforms), the IITP grant (IITP-2026-RS-2026-25615334, AI Star Fellowship Support Program), and an IITP grant for the AI Computing Support Project for R\&D (RS-2026-25505492, High-Performance Research AI Computing Infrastructure Support at the 2 PFLOPS Scale), all funded by the Ministry of Science and ICT (MSIT), Republic of Korea.


%
%
\bibliographystyle{splncs04}
\bibliography{main}
\end{document}